\documentclass[11pt]{article}

\usepackage[preprint]{acl}

\usepackage{times}
\usepackage{latexsym}
\usepackage[T1]{fontenc}
\usepackage[utf8]{inputenc}
\usepackage{microtype}
\usepackage{inconsolata}
\usepackage{graphicx}

\usepackage{amsmath,amssymb,amsfonts}
\usepackage{booktabs}
\usepackage{array}
\newcolumntype{L}[1]{>{\raggedright\arraybackslash}p{#1}}
\usepackage{url}
\usepackage{seqsplit}  
\usepackage{listings}  
\usepackage{tcolorbox}
\tcbuselibrary{listings,breakable,skins}
\newtcblisting{promptbox}[2][]{
    enhanced,
    breakable,
    listing only,
    title={#2},
    title after break={#2\space (continued)},
    colback=black!3,
    colframe=black!70,
    colbacktitle=black!75,
    coltitle=white,
    fonttitle=\bfseries\footnotesize,
    boxrule=0.5pt,
    arc=1.5mm,
    left=5pt, right=5pt, top=5pt, bottom=5pt,
    before skip=8pt, after skip=8pt,
    listing options={
        basicstyle=\ttfamily\footnotesize,
        breaklines=true,
        breakindent=0pt,
        breakatwhitespace=false,
        columns=fullflexible,
        keepspaces=true,
        showstringspaces=false,
        alsoother={_},
        frame=none,
        aboveskip=0pt, belowskip=0pt,
        postbreak={}
    },
    #1
}

\title{POS-Signature Demonstration Retrieval for In-Context Dependency Parsing}

\author{
  \textbf{Muhammad Azam Afridi\textsuperscript{1,2,3}} \quad
  \textbf{Rajab Ali\textsuperscript{4}} \quad
  \textbf{Marco Passarotti\textsuperscript{1}} \\
  \textsuperscript{1}Universit\`a Cattolica del Sacro Cuore, Milan, Italy \\
  \textsuperscript{2}University of Zaragoza, Zaragoza, Spain \\
  \textsuperscript{3}Universidade NOVA de Lisboa, Lisbon, Portugal \\
  \textsuperscript{4}Universit\'e libre de Bruxelles, Brussels, Belgium \\
  \small{\texttt{azamafridi.ai@gmail.com} \quad
  \texttt{rajab.ali@ulb.be} \quad
  \texttt{marco.passarotti@unicatt.it}}
}

\begin{document}
\maketitle

\begin{abstract}
Demonstrations for in-context learning are often retrieved by semantic similarity, which may be a poor proxy for relevance in structured prediction. We study demonstration selection for Universal Dependencies parsing with a POS-signature policy that represents sentences with weighted part-of-speech statistics and retrieves three length-compatible, gold-parsed examples from the labeled training split. We evaluate two open-weight language models without fine-tuning on all 2,077 sentences of the UD English Web Treebank test set, comparing the policy with shot- and format-matched fixed and semantic three-shot baselines, as well as zero-shot, chain-of-thought, and critique-refinement systems. POS-signature retrieval yields the highest labeled attachment score among the tested prompting systems for both models (72.4 and 73.2), outperforming semantic retrieval by 1.9 and 2.7 points; paired-bootstrap 95\% confidence intervals exclude zero for both contrasts. Chain-of-thought and refinement comparisons vary across models. These exploratory results favor the tested POS-signature retrieval approach on English-EWT. We release the prompts, archived outputs, and evaluation code.
\end{abstract}

\section{Introduction}

Retrieval-based in-context learning commonly selects demonstrations whose text is semantically similar to the query \citep{liu2022}. Demonstration selection can be viewed as a form of context engineering: selecting and organizing the information supplied to an LLM at inference time \citep{mei2025}. Semantic proximity is useful for topic-sensitive tasks, but it may be a poor proxy for relevance when outputs depend on structure: sentences about different topics can share a POS pattern, while topically similar sentences can require different structural analyses.

Dependency parsing makes this distinction concrete. Given a tokenized sentence, a parser assigns every token a syntactic head and dependency relation, producing a rooted tree under an annotation scheme such as Universal Dependencies (UD; \citealp{nivre2016}). Large language models can emit dependency and constituency structures, but their accuracy depends strongly on prompt design, demonstrations, and task decomposition \citep{tian2024,ezquerro2025,matsuda2025,ginn2025}. We ask whether a POS-signature retrieval policy produces more accurate parses than general semantic retrieval under the same prompt scaffold and three-example budget.

Our policy uses a supervised Stanza tagger to predict the query's POS sequence and represents it through weighted features of sentence length, POS composition, and adjacent POS transitions. It first restricts the labeled English-EWT training pool by length compatibility, then retrieves the three signatures with highest cosine similarity and inserts their gold parses as demonstrations. Predicted query tags control retrieval only and are not shown to the language model. Because the semantic baseline searches the full training pool, our central comparison evaluates complete retrieval policies rather than isolating the representation itself.

We compare six prompting conditions on all 2,077 English-EWT test sentences using OpenAI's \texttt{gpt-oss-120b} and Qwen's \texttt{Qwen2.5-72B-Instruct}, without fine-tuning. Fixed selection uses the same three training examples for every query, whereas POS-signature and semantic retrieval select query-specific examples using POS-pattern similarity with length filtering and sentence-embedding similarity, respectively. Fixed, POS-signature, and semantic few-shot conditions are shot- and format-matched; zero-shot, chain-of-thought (CoT), and critique-refinement provide broader system comparisons. The primary metric is token-weighted labeled attachment score (LAS) over 21,998 non-punctuation tokens, with paired sentence-bootstrap inference for designated contrasts.

POS-signature retrieval achieves the highest LAS among the tested prompting systems for both models; all four primary retrieval contrasts remain significant after Holm correction. CoT instead improves over zero-shot for gpt-oss but reduces accuracy for Qwen, showing a model-dependent difference between these bundled systems.

This paper makes three contributions:
\begin{enumerate}
    \item \textbf{POS-signature demonstration retrieval:} an interpretable policy that selects three gold-parsed examples using explicit POS statistics and length compatibility.
    \item \textbf{Cross-model policy comparison:} six conditions over the full English-EWT test split, including three shot- and format-matched demonstration policies.
    \item \textbf{Evidence and diagnostics:} exploratory retrieval-policy results across two models, alongside analyses of relation-level accuracy and changes introduced by refinement. We release all prompts, archived outputs, evaluation code, and reproduction instructions.\footnote{\url{https://github.com/azamafridi23/pos-signature-parsing}; archived version 0.1.0: \url{https://doi.org/10.5281/zenodo.22924472}.}
\end{enumerate}

The reported LLM results can be reproduced from the archived outputs without additional model API calls. The findings are scoped to two hosted models, one English treebank, gold tokenization, and labeled in-domain demonstrations.

\section{Related Work}
\label{sec:related-work}

Prompted syntactic parsing remains challenging: many open LLMs fail to beat input-agnostic baselines in multilingual zero-shot evaluation \citep{ezquerro2025}. More explicit guidance can help, including intermediate reasoning for constituency parsing \citep{tian2024}, POS-first task decomposition for dependency parsing \citep{matsuda2025}, and rule-based instructions, which improve zero-shot dependency parsing but lose their benefit when a few labeled demonstrations are provided \citep{ginn2025}. Closest to our setting, \citet{adel2026} retrieve Arabic examples for morphosyntactic tagging and dependency parsing using chrF++ overlap or sentence-embedding similarity, with the demonstration count tuned on development data. We instead retrieve English UD demonstrations by POS signature, fix \(k=3\), and compare against a matched sentence-embedding policy.

Demonstration selection typically favors semantic similarity \citep{liu2022}, but task-specific structure has guided selection for semantic parsing and machine translation. \citet{gupta2023} evaluate dependency-subtree coverage among several coverage-based selectors for semantic parsing, and \citet{tang2024} combine syntactic and lexical coverage for translation. Our setting differs in prediction target and representation: we use sentence-level POS statistics to retrieve examples for dependency-tree prediction and compare the resulting policy with fixed selection and Sentence-BERT retrieval \citep{reimers2019}, built on MPNet \citep{song2020}, while holding demonstration count and serialization constant.

CoT and iterative refinement provide alternative prompting interventions. CoT is effective for some reasoning tasks but less consistent elsewhere \citep{wei2022,sprague2025}; self-refinement can likewise repair or introduce errors without external feedback \citep{madaan2023,huang2024}. We therefore evaluate CoT and critique-refinement as broader prompting systems, not factor-matched retrieval baselines.

\section{Method}
\label{sec:method}

\subsection{Task and Resource Setting}
\label{sec:task-setting}

Let a tokenized sentence be \(x=(x_1,\ldots,x_n)\). For each token \(x_i\), the model predicts a head \(h_i \in \{0,\ldots,n\}\), where 0 denotes the root, and a UD relation \(r_i\). Although the prompt also requests a universal POS tag, evaluation scores heads and relations. All first-pass queries provide gold-tokenized \texttt{(id, form)} pairs but withhold the query's gold POS tags, lemmas, morphological features, heads, and dependency relations.

The five first-pass conditions share an instruction defining the task, UD inventories, root convention, and JSON schema, followed by condition-specific demonstrations or reasoning instructions. Critique-Refine separately reviews a CoT parse. Model parameters remain frozen, but demonstrations contain complete gold parses from the labeled EWT training split; the setting is therefore supervised in-context parsing, not zero-resource parsing.

\subsection{Prompting Conditions}
\label{sec:prompting-conditions}

We evaluate six prompting conditions (Table~\ref{tab:prompting-conditions}). Fixed, POS-Signature, and Semantic Few-Shot share the task instruction, three-example budget (\(k=3\)), and demonstration format, enabling direct comparison of their selection policies. CoT and Critique-Refine also change the example count, response format, or number of calls and are therefore system-level comparisons.

\begin{table*}[t]
\centering
\footnotesize
\begin{tabular}{lclcc}
\toprule
\textbf{Condition} & \textbf{Demonstrations} & \textbf{Selection} & \textbf{Free-text reasoning} & \textbf{Passes} \\
\midrule
Zero-Shot & 0 & --- & No & 1 \\
Fixed Few-Shot & 3 & fixed, manually selected & No & 1 \\
POS-Signature Few-Shot & 3 & POS signature + length window & No & 1 \\
Semantic Few-Shot & 3 & full-pool sentence embeddings & No & 1 \\
Chain-of-Thought & 1 & fixed worked example & Yes & 1 \\
Critique-Refine & --- & revises CoT output & Internal review & 2 \\
\bottomrule
\end{tabular}
\caption{Prompting conditions. The three few-shot policies are shot- and format-matched; comparisons involving CoT or Critique-Refine are system-level.}
\label{tab:prompting-conditions}
\end{table*}

Zero-Shot uses the shared instruction alone. Fixed Few-Shot uses the same three manually selected EWT training examples for every query. POS-Signature and Semantic Few-Shot instead retrieve three query-specific examples using the policies in Sections~\ref{sec:syntax-retrieval} and~\ref{sec:semantic-retrieval}. CoT provides one worked EWT example and permits free-text reasoning before the JSON parse. Critique-Refine makes a dependent second call that reviews the extracted CoT parse without its rationale or any gold annotation; if extraction fails, the second call is omitted and the sentence receives zero credit under the common failure policy. Appendix~\ref{app:prompts} presents the prompt templates; complete instructions and demonstrations are available in the repository.

\subsection{POS-Signature Demonstration Retrieval}
\label{sec:syntax-retrieval}

\begin{figure*}[t]
\centering
\IfFileExists{figures/fig_pos_signature_retrieval.pdf}{%
    \includegraphics[width=\textwidth]{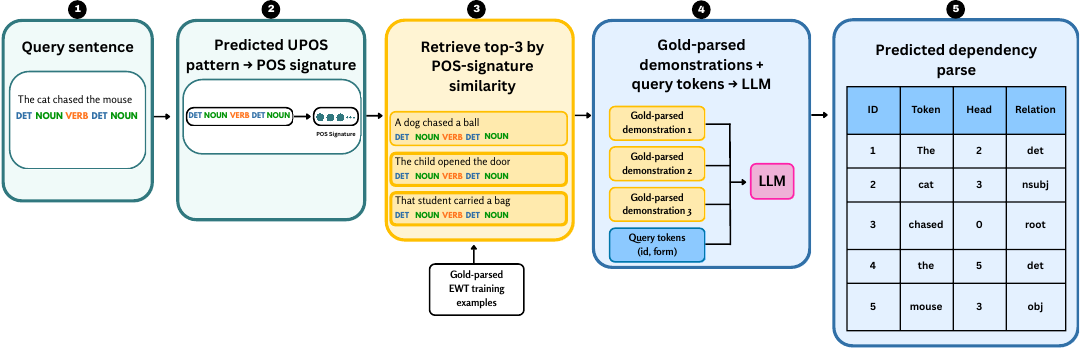}%
}{%
    \fbox{\parbox[c][1.25in][c]{0.92\textwidth}{\centering
    Figure asset not found: \texttt{figures/fig\_pos\_signature\_retrieval.pdf}}}%
}
\caption{POS-Signature retrieval. Predicted query UPOS selects three length-compatible, gold-parsed demonstrations but is not shown to the LLM; sentences are illustrative.}
\label{fig:retrieval-pipeline}
\end{figure*}

POS-Signature retrieval selects demonstrations using POS composition and local POS order (Figure~\ref{fig:retrieval-pipeline}). Its retrieval features require POS tags but no dependency arcs, lemmas, or morphological features; gold dependency parses supply the demonstration outputs.

\paragraph{Sentence representation.} For sentence \(s\), let \(z(s)=(z_1,\ldots,z_n)\) be its UPOS sequence, including punctuation. Let \(c_a\) and \(c_{ab}\) count tag \(a\) and adjacent tag pair \((a,b)\), respectively. The EWT training pool fixes vocabularies \(V_u\) of 17 tags and \(V_b\) of 280 observed bigrams. We form four feature blocks using scaled length and normalized frequencies:
\begin{equation}
\begin{aligned}
f_{\ell} &= [\,n/(n{+}10),\; c_{\mathrm{PUNCT}}/n, \\
&\quad (c_{\mathrm{VERB}}{+}c_{\mathrm{AUX}})/n, \\
&\quad c_{\mathrm{CCONJ}}/n,\; c_{\mathrm{SCONJ}}/n\,], \\
f_u &= [\,c_a/n\,]_{a\in V_u}, \\
f_b &= [\,c_{ab}/\max(1,n{-}1)\,]_{(a,b)\in V_b}, \\
f_r &= [\,c_{\mathrm{NOUN}}/n,\; (c_{\mathrm{VERB}}{+}c_{\mathrm{AUX}})/n, \\
&\quad c_{\mathrm{ADP}}/n,\; c_{\mathrm{PRON}}/n\,].
\end{aligned}
\label{eq:feature-blocks}
\end{equation}
Here \(f_\ell\) encodes length and coarse composition, \(f_u\) tag frequencies, \(f_b\) local POS order, and \(f_r\) selected category ratios derived from the same counts as \(f_u\). The concatenation has \(5+17+280+4=306\) dimensions.

\paragraph{Weighting and normalization.} We multiply the blocks in \eqref{eq:feature-blocks} by fixed coefficients, concatenate them, and apply one global \(\ell_2\) normalization:
\begin{equation}
\begin{split}
\phi(s) = \operatorname{norm}_2\big( [\,&0.2f_{\ell};\; 0.3f_u; \\
&0.4f_b;\; 0.1f_r\,] \big).
\end{split}
\label{eq:signature}
\end{equation}
The blocks are not separately \(\ell_2\)-normalized. The multipliers are hand specified, not learned or optimized, and do not represent fixed shares of cosine similarity: each block's contribution also depends on its feature values and norm. Global normalization makes the dot product between signatures equal cosine similarity.

\paragraph{Candidate filtering and selection.} We first retain a training sentence \(t\) when
\begin{equation}
\left|\,|t|-|s|\,\right| \leq
\max\!\left(3,\left\lfloor 0.3|s|\right\rfloor\right).
\label{eq:length-window}
\end{equation}
Thus the allowable length difference is at least three tokens and grows with the query; lengths are those of the UPOS sequences used by the retriever. If this candidate pool is empty, we use the full training pool. Candidate similarity is
\begin{equation}
\operatorname{sim}(s,t)=\phi(s)^{\top}\phi(t),
\label{eq:syntax-similarity}
\end{equation}
Candidates are ranked by decreasing similarity, skipping exact query-text matches and duplicate selected texts. The first \(k=3\) remaining candidates are inserted with their gold token records (POS tags, heads, and relations); every archived test prompt received three demonstrations.

\paragraph{Tagging provenance.} Training signatures use the gold UPOS sequences of the 12,544 EWT training sentences. Query signatures use the UPOS sequence predicted from sentence text by Stanza's \texttt{tokenize,pos} pipeline \citep{qi2020}. Gold test UPOS is never used for retrieval, and predicted query tags guide selection only: they are not placed in the prompt or otherwise exposed to the LLM.

\subsection{Semantic Retrieval Baseline}
\label{sec:semantic-retrieval}

Semantic Few-Shot encodes each raw sentence with the frozen \texttt{all-mpnet-base-v2} Sentence-Transformer \citep{sentence-transformers2025}, using its configured mean pooling to produce a 768-dimensional vector. We precompute vectors for all 12,544 training sentences and \(\ell_2\)-normalize training and query embeddings, making their dot product cosine similarity. The repository documents the encoder revision, sorting behavior, and environment metadata for both retrievers.

An exhaustive full-pool search ranks candidates by decreasing similarity, applies the same exact-text exclusion and deduplication, and selects three demonstrations in the shared format. Unlike POS-Signature retrieval, this policy has no length filter. The comparison therefore varies both representation and candidate pool (Appendix~\ref{app:factor-audit}).

\section{Experimental Setup}
\label{sec:experimental-setup}

\subsection{Data}
\label{sec:data}

We use UD English Web Treebank v2.17 \citep{ud2025,silveira2014}. Its training, development, and test splits contain 12,544, 2,001, and 2,077 sentences; training supplies the in-domain demonstration pool, development is unused, and all test sentences are evaluated. We retain integer-ID CoNLL-U word rows, including punctuation, but exclude multiword-token ranges and empty nodes, with no other corpus preprocessing. Of 25,094 test tokens, 3,096 gold \texttt{PUNCT} tokens are excluded, leaving a fixed scoring denominator of 21,998.

Models receive gold-tokenized \texttt{(id, form)} pairs, isolating dependency parsing from segmentation and making the task narrower than end-to-end parsing. POS-Signature retrieval additionally uses Stanza-predicted query UPOS, which is never shown to the LLM; gold test UPOS, heads, and relations are used only for evaluation.

\subsection{Models and Inference Configuration}
\label{sec:models-inference}

We evaluate two open-weight models without fine-tuning: OpenAI's reasoning model \texttt{gpt-oss-120b} \citep{openai2025} and Qwen's dense instruction-tuned \texttt{Qwen2.5-72B-\allowbreak Instruct} \citep{qwen2024}. Both models are served via DeepInfra's OpenAI-compatible chat API \citep{deepinfra-chat2026}. We use the provider's documented gpt-oss and Qwen endpoints \citep{deepinfra-gptoss2026,deepinfra-qwen2026}. The systems differ in developer and post-training regime but do not represent the broader LLM landscape.

All conditions use temperature 0, top-\(p=1\), seed 42, and a 6,000-token completion limit; every first-pass condition except CoT requests a JSON-object response. These settings reduce sampling variation but do not make hosted inference deterministic. gpt-oss uses reasoning effort \texttt{none}; Qwen omits the reasoning-effort field. The repository documents retry behavior and archived endpoint metadata.

\subsection{Metrics and Statistical Analysis}
\label{sec:metrics}

Responses are decoded against the requested JSON schema (from the first \texttt{\{} for CoT) and aligned to gold tokens by identifier. Extraction failures and missing identifiers receive zero credit; predictions need not form a well-formed tree. Decoding rejects trailing non-whitespace text. Appendix~\ref{app:tree-audit} describes the separate structural checks.

We report standard unlabeled and labeled attachment scores (UAS/LAS; \citealp{zeman2018}) over a fixed denominator of \(N=21{,}998\) non-punctuation gold tokens, with LAS primary and UAS secondary. Gold UPOS identifies \texttt{PUNCT}; full relation strings, including subtypes, are compared case-insensitively. Failed or missing predictions and absent Critique-Refine records remain in \(N\) with zero correct arcs, while structurally ill-formed outputs remain token-level scorable. Because this fixed-denominator, soft-validation protocol also uses gold tokenization and retains relation subtypes, it differs from official CoNLL 2018 scoring \citep{conll18-eval}; scores are directly comparable only under matching choices. Appendix~\ref{app:metric-definitions} gives the exact formulas.

For each contrast, we pair the same 2,077 sentences and draw 10,000 sentence-bootstrap replicates, applying identical resampled indices to both conditions and recomputing token-weighted LAS \citep{koehn2004}. We report the observed percentage-point difference, percentile 95\% confidence interval, and approximate two-sided bootstrap p-value; Appendix~\ref{app:metric-definitions} defines the calculation.

We apply Holm correction at \(\alpha=.05\) jointly to four focal comparisons: POS-Signature versus Semantic and versus Fixed for each model \citep{holm1979}. Other comparisons are exploratory; Appendix~\ref{app:results} reports all 15 pairwise comparisons per model.

\subsection{Supervised Reference}
\label{sec:supervised-reference}

For context, we evaluate Stanza 1.11.1, a supervised biaffine parser \citep{qi2020,dozat2017}, on the same gold-tokenized test set with the same evaluator, punctuation exclusion, and 21,998-token denominator. It scores 90.7 UAS and 88.0 LAS. Because it is task-supervised, Stanza is a contextual reference rather than a resource-matched prompting baseline.

\section{Results}
\label{sec:results}

\subsection{Overall Performance and Primary Contrasts}
\label{sec:overall-results}

POS-Signature Few-Shot leads all prompting conditions for both models (Table~\ref{tab:overall-results}), reaching 80.4 UAS/72.4 LAS on gpt-oss and 82.1 UAS/73.2 LAS on Qwen. Semantic Few-Shot reaches 70.5 LAS on both. The supervised Stanza reference remains substantially higher at 88.0 LAS, gaps of 15.6 and 14.8 points from the strongest prompted systems.

\begin{table*}[t]
\centering
\footnotesize
\begin{tabular}{lrrrr}
\toprule
\textbf{Condition} & \multicolumn{2}{c}{\textbf{gpt-oss}} & \multicolumn{2}{c}{\textbf{Qwen}} \\
\cmidrule(lr){2-3}\cmidrule(lr){4-5}
 & \textbf{UAS (\%)} & \textbf{LAS (\%)} & \textbf{UAS (\%)} & \textbf{LAS (\%)} \\
\midrule
Zero-Shot & 75.1 & 64.7 & 75.7 & 62.9 \\
Fixed Few-Shot & 76.5 & 67.3 & 76.3 & 65.3 \\
\textbf{POS-Signature Few-Shot} & \textbf{80.4} & \textbf{72.4} & \textbf{82.1} & \textbf{73.2} \\
Semantic Few-Shot & \underline{79.6} & \underline{70.5} & \underline{80.8} & \underline{70.5} \\
Chain-of-Thought & 78.6 & 70.0 & 72.4 & 61.7 \\
Critique-Refine & 79.1 & 70.4 & 74.6 & 63.6 \\
\midrule
\emph{Stanza supervised reference} & \emph{90.7} & \emph{88.0} & \emph{90.7} & \emph{88.0} \\
\bottomrule
\end{tabular}
\caption{Corpus UAS/LAS (\%) on 2,077 EWT test sentences and 21,998 non-punctuation tokens; unscorable outputs receive zero. Bold and underline mark the best and second-best LLM results, not significance. The duplicated Stanza values represent one reference run.}
\label{tab:overall-results}
\end{table*}

The paired-bootstrap analysis supports the retrieval advantage (Table~\ref{tab:primary-contrasts}). Relative to Semantic Few-Shot, POS-Signature improves LAS by 1.87 points on gpt-oss (95\% CI [1.06, 2.69]) and 2.68 on Qwen (95\% CI [1.97, 3.41]); gains over Fixed Few-Shot are 5.09 and 7.86 points. The corresponding UAS gains over Semantic Few-Shot are smaller, at 0.8 and 1.3 points, suggesting that improved relation labeling contributes substantially to the LAS advantage. All four Holm-adjusted \(p\)-values are .0008. These estimates compare complete policies because representation and candidate filtering both differ.

\subsection{CoT and Refinement System Comparisons}
\label{sec:reasoning-results}

CoT reverses direction across models: relative to Zero-Shot, it gains 5.30 LAS points on gpt-oss (95\% CI [4.41, 6.19]) but loses 1.18 on Qwen (95\% CI [$-2.22$, $-0.21$]). It therefore approaches the stronger few-shot conditions on gpt-oss but ranks last on Qwen.

\begin{table*}[t]
\centering
\footnotesize
\begin{tabular}{llrlrl}
\toprule
\textbf{Comparison (first $-$ second)} & \textbf{Status} & \multicolumn{2}{c}{\textbf{gpt-oss}} & \multicolumn{2}{c}{\textbf{Qwen}} \\
\cmidrule(lr){3-4}\cmidrule(lr){5-6}
 & & \textbf{$\Delta$LAS pp [95\% CI]} & \textbf{$p$} & \textbf{$\Delta$LAS pp [95\% CI]} & \textbf{$p$} \\
\midrule
POS-Signature $-$ Semantic & Primary & $+1.87$ [1.06, 2.69] & $.0008^{\dagger}$ & $+2.68$ [1.97, 3.41] & $.0008^{\dagger}$ \\
POS-Signature $-$ Fixed & Primary & $+5.09$ [4.21, 5.98] & $.0008^{\dagger}$ & $+7.86$ [7.05, 8.71] & $.0008^{\dagger}$ \\
CoT $-$ Zero-Shot & Exploratory & $+5.30$ [4.41, 6.19] & .0002 & $-1.18$ [$-2.22$, $-0.21$] & .0158 \\
Critique-Refine $-$ CoT & Exploratory & $+0.44$ [$-0.02$, 0.91] & .0634 & $+1.90$ [1.53, 2.28] & .0002 \\
\bottomrule
\end{tabular}
\caption{Paired-bootstrap LAS differences (pp) with percentile 95\% CIs from 10,000 paired sentence resamples. \(\dagger\) Primary \(p\)-values are Holm-adjusted across four tests; exploratory values are unadjusted. All tests are two-sided; Appendix~\ref{app:paired-comparisons} reports the remaining comparisons.}
\label{tab:primary-contrasts}
\end{table*}

Critique-Refine gains 0.44 points over CoT on gpt-oss (95\% CI [$-0.02$, 0.91]) and 1.90 on Qwen (95\% CI [1.53, 2.28]). For Qwen Critique-Refine minus Zero-Shot, the difference is $+0.72$ points (95\% CI [$-0.28$, $+1.64$], \(p=.1484\); Table~\ref{tab:appendix-comparisons}); the interval includes zero, so no statistically clear difference was detected. These system comparisons depend on the available first-pass parses; three Qwen second-pass records are absent because their CoT outputs contain no scorable JSON.

\subsection{Output Compliance}
\label{sec:output-reliability}

Schema-valid output rates are 99.7--100\%. All evaluated gpt-oss responses are schema-valid; Qwen failures comprise six Zero-Shot, one Fixed, one POS-Signature, and three truncated CoT responses. The CoT failures also prevent their Critique-Refine calls, while Semantic Few-Shot has none. Appendix~\ref{app:tree-audit} gives full counts.

Root warnings occur in at most 14 responses per condition and remain token-level scorable. Output compliance and attachment accuracy are distinct measures. Appendix~\ref{app:tree-audit} reports a stricter, non-scoring tree audit.

\subsection{Post-Hoc Diagnostic Analyses}
\label{sec:diagnostics}

These post-hoc diagnostics localize aggregate differences by relation, sentence length, and refinement behavior; they do not establish causal mechanisms.

\subsubsection{Relation-Level Localization}
\label{sec:relation-localization}

POS-Signature exceeds Semantic retrieval under both models for \texttt{nmod:poss}, \texttt{cop}, and \texttt{nmod:unmarked}. The largest gap is for \texttt{nmod:poss}: 13.9 LAS points on gpt-oss and 23.0 on Qwen, although other example-based conditions also improve this relation. Only the retrieval policies obtain nonzero LAS on \texttt{nmod:unmarked}, whose 111-token support warrants caution. Appendix~\ref{app:relation-results} reports all six diagnostic relations. These patterns do not identify whether structural match, relation coverage, or lexical overlap causes the gains.

\subsubsection{Sentence Length and Refinement Behavior}
\label{sec:length-refinement}

Critique-Refine is numerically above POS-Signature for gpt-oss in the 21--30 and 31--50 token bins, whereas POS-Signature leads every Qwen bin (Appendix~\ref{app:length-results}). Refinement changes similar token shares (10.3\% and 9.6\%) but yields net gains of 97 LAS-correct tokens for gpt-oss and 418 for Qwen; Appendix~\ref{app:refinement-edits} separates fixes from regressions.

\section{Discussion}
\label{sec:discussion}

\subsection{Interpreting the Retrieval-Policy Advantage}
\label{sec:retrieval-discussion}

The retrieval-policy advantage can be interpreted from two complementary perspectives: linguistic structure and how demonstrations guide the LLM's predictions.

From a linguistic perspective, POS composition, adjacent POS transitions, and sentence length capture surface patterns that may help identify relevant constructions. Such patterns may be particularly useful in English, where constituent order and function-word placement constrain dependency relations. Previous work supports the relevance of POS-sequence statistics and word-order properties to dependency parsing \citep{wang2018,gulordava2016}. However, POS signatures summarize tag composition and adjacent transitions but omit head attachments and dependency labels; similar signatures can therefore correspond to different dependency structures.

From an in-context learning perspective, relevant demonstrations may help the model apply UD head and relation-label conventions to similar constructions. The larger LAS than UAS advantage is consistent with this explanation. For example, improvements in \texttt{nmod:poss} and \texttt{cop} may reflect guidance on possessive constructions and UD's predicate-headed copular analysis \citep{demarneffe2021}. Nevertheless, \texttt{nmod:unmarked} remains difficult, and Semantic Few-Shot performs better for Qwen on \texttt{obl:unmarked}, suggesting a potential complementary role for lexical-semantic information.

These interpretations remain hypotheses. The comparison evaluates the complete retrieval policy and does not isolate POS features from length filtering, predicted-POS errors, or incidental lexical overlap. The observed advantage is therefore specific to the tested models and data.

\subsection{Interpreting CoT and Refinement Comparisons}
\label{sec:reasoning-discussion}

The CoT sign reversal concerns a bundled prompting system, not an isolated ``reason step by step'' instruction. gpt-oss and Qwen differ in architecture, scale, training, tokenization, serving, and reasoning controls, and their CoT outputs differ in length and failure behavior. With one model per family, the experiment cannot identify which factor explains the reversal.

Refinement can repair errors or overwrite correct arcs. Its gain over CoT is uncertain for gpt-oss; for Qwen, it offsets CoT's loss, but the contrast with Zero-Shot is also uncertain (Section~\ref{sec:reasoning-results}). These comparisons do not establish that critique or first-pass weakness causes improvement. No interaction test was specified for the length crossover, and the longest bin contains only 11 scored sentences.

\subsection{Implications for Evaluating LLM Parsers}
\label{sec:parsing-implications}

Schema compliance is not parsing success. Although 99.7--100\% of outputs are schema-valid, the post-hoc audit finds tree-conforming rates of 96.5--98.4\% for gpt-oss and 96.9--99.1\% for Qwen, while the best LAS is only 72--73\% (Appendix~\ref{app:tree-audit}). Moreover, gpt-oss POS-Signature Few-Shot has the highest LAS but the lowest tree-conforming count, showing that better attachment and labeling need not enforce global well-formedness. Evaluations should therefore report serialization reliability, structural validity, and parsing accuracy separately.

\section{Conclusion}
\label{sec:conclusion}

This study examines how demonstration-retrieval policies affect dependency parsing by frozen LLMs. Among six prompting systems, POS-Signature retrieval achieves the highest LAS for both models on English-EWT, exceeding full-pool semantic retrieval by 1.9 and 2.7 points. CoT reverses direction across models, and refinement produces both repairs and regressions.

These exploratory results favor the tested POS-signature retrieval approach, which requires a supervised tagger and labeled demonstration pool. They do not isolate POS features from length filtering or establish an alternative to supervised parsing; matched ablations remain needed. Future work could investigate development-set tuning of feature weights, alternative measures of structural similarity for demonstration selection, and the effects of the sentence-length window and number of retrieved examples.

\section*{Limitations}

The retrieval comparison evaluates complete policies: POS-Signature combines POS features with length filtering, whereas Semantic retrieval searches the full training pool. It therefore does not isolate representation from candidate filtering, and the retrieval components and hyperparameters were not ablated. Although the LLMs are frozen, the approach relies on a supervised tagger and gold-annotated demonstrations from the evaluation treebank's training split, limiting claims about low-resource use. Its scores remain below the supervised Stanza reference. CoT and refinement also change the example count, response format, or number of calls, so these comparisons do not isolate the effects of reasoning or critique.

The findings cover only two models and English-EWT with gold sentence boundaries and tokens. Differences in model training and serving prevent attributing their contrasting CoT results to a single factor. English's dominant constituent order and limited inflectional morphology may make local POS patterns informative in ways that do not transfer to languages with freer word order or richer case morphology \citep{dehouck2018}. Generalization to other languages and domains, and to end-to-end parsing with segmentation and tokenization errors, remains untested.

Prompts, retrieval weights, and fixed examples were finalized before evaluation; test outputs were inspected afterward to interpret the results. Relation- and length-based diagnostics are therefore post-hoc and should be interpreted as exploratory. Unknown pretraining exposure to the public EWT data and UD guidelines cannot be excluded. Finally, token-level attachment scores do not guarantee well-formed dependency trees or establish downstream utility; structural validity is assessed separately.

\section*{Ethical Considerations}

This study uses the public UD English-EWT dataset. Its source text may contain identifying information; inputs and demonstrations were processed through DeepInfra and included in the released archives without additional de-identification. The artifact retains the dataset's license and source notices \citep{ud-ewt-license2025}. Generated parses are model predictions and should not be treated as verified annotations.

\section*{Acknowledgments}

ChatGPT (OpenAI) was used for language editing and improving the readability of the manuscript. The authors reviewed and take full responsibility for the final content.

\bibliography{custom}

\appendix

\setcounter{dbltopnumber}{2}

\section{Prompt Templates}
\label{app:prompts}

This appendix presents the prompt templates used in the experiments. Complete instructions, fixed demonstrations, and archived messages are available in the accompanying repository. POS-Signature and Semantic retrieval share the same base instruction and demonstration format; only their selected examples differ.

\subsection{Shared Task Instruction}
\label{app:shared-instruction}

The following is the exact Zero-Shot system message. It also provides the common task definition for the other first-pass conditions. Box titles are presentation labels; line wrapping is for readability.

\begin{nolinenumbers}
\begin{promptbox}{System message: Zero-Shot}
You are a Universal Dependencies (UD) syntactic parser for English.
You will be given an English sentence and a numbered list of its tokens
(id and word form).

Your task is to produce a UD parse in JSON with this exact structure:
{
  "tokens": [
    {
      "id":     <integer, matches the input token id>,
      "form":   "<copy the word form exactly from input>",
      "upos":   "<Universal POS tag>",
      "head":   <integer, id of syntactic head; 0 means this token is root>,
      "deprel": "<UD dependency relation label>"
    },
    ...
  ]
}

Rules:
- Output ONLY the raw JSON object. No markdown, no explanation.
- Exactly one token must have head=0 AND deprel="root". No other token
  may use head=0 or deprel="root".
- Use UD v2 UPOS tags: ADJ, ADP, ADV, AUX, CCONJ, DET, INTJ, NOUN,
  NUM, PART, PRON, PROPN, PUNCT, SCONJ, SYM, VERB, X
- Use standard UD v2 deprel labels: acl, advcl, advmod, amod, appos,
  aux, case, cc, ccomp, compound, conj, cop, csubj, dep, det, discourse,
  dislocated, expl, fixed, flat, goeswith, iobj, list, mark, nmod,
  nsubj, nummod, obj, obl, orphan, parataxis, punct, reparandum, root,
  vocative, xcomp
- Subtypes are allowed using colon notation (e.g. nsubj:pass, aux:pass,
  acl:relcl, obl:agent). Use them where appropriate.
- Copy id and form exactly as given; do not add or remove tokens.
\end{promptbox}
\end{nolinenumbers}

Fixed Few-Shot adds three fixed gold-parsed examples. Both retrieval conditions use a slightly extended instruction and three selected examples. CoT provides one worked example and allows reasoning before the final JSON. Complete versions are preserved in \path{docs/prompt_source_snapshot.md}.

\subsection{Query and Demonstration Templates}
\label{app:dynamic-messages}

All five first-pass conditions use this query format:

\begin{nolinenumbers}
\begin{promptbox}[unbreakable]{User message: first pass}
# text = {sentence_text}
{token_id_1}\t{token_form_1}
...
{token_id_n}\t{token_form_n}
\end{promptbox}
\end{nolinenumbers}

Both retrieval conditions append three demonstrations in the following format:

\begin{nolinenumbers}
\begin{promptbox}[unbreakable]{Retrieved demonstration template}
Input:
# text = {retrieved_training_sentence}
{token_id_1}\t{token_form_1}
...
{token_id_n}\t{token_form_n}

Output:
{
  "tokens": [
    {
      "id": {integer},
      "form": "{exact_form}",
      "upos": "{gold_upos}",
      "head": {gold_head},
      "deprel": "{gold_deprel}"
    }
  ]
}
\end{promptbox}
\end{nolinenumbers}

Braced placeholders are filled with the selected training data. Both retrieval conditions use the same formatting.

\subsection{Critique-Refine Messages}
\label{app:critique-messages}

Critique-Refine reviews the extracted CoT parse. Its full system instruction is preserved in the repository; the standard user-message template is shown below.

\begin{nolinenumbers}
\begin{promptbox}[unbreakable]{User message: Critique-Refine}
Sentence: {sentence_text}

Tokens:
{token_id_1}\t{token_form_1}
...
{token_id_n}\t{token_form_n}

Existing parse:
{
  "tokens": {extracted_CoT_token_array_serialized_with_indent_2}
}
\end{promptbox}
\end{nolinenumbers}

\section{Retrieval-Policy Comparison}
\label{app:factor-audit}

POS-Signature represents sentences using weighted POS statistics and filters candidates by length; Semantic retrieval uses sentence embeddings and searches the full training pool. Both rank candidates by cosine similarity, exclude exact query-text matches and duplicate selections, and supply three gold-parsed examples in the same format under matched inference settings. Their comparison therefore evaluates complete retrieval policies, not POS representation alone. Sections~\ref{sec:syntax-retrieval}--\ref{sec:semantic-retrieval} describe the policies.

\section{Additional Results}
\label{app:results}

All corpus-level comparisons use the same 2,077 sentences and 21,998 non-punctuation gold tokens. Missing or unscorable predictions receive zero credit without reducing the denominator. Exceptions for the refinement-transition analysis are stated below.

\subsection{Scoring and Statistical Tests}
\label{app:metric-definitions}

Let \(N=21{,}998\), with predicted heads and labels \(h_i,r_i\) and gold values \(h_i^*,r_i^*\). UAS requires the correct head; LAS requires both the correct head and the full relation label, including any subtype. Label comparison is case-insensitive.

\begin{equation}
\begin{aligned}
\operatorname{UAS} &= 100\frac{\sum_{i=1}^{N}\mathbf{1}[h_i=h_i^\ast]}{N}, \\
\operatorname{LAS} &= 100\frac{\sum_{i=1}^{N}
\mathbf{1}[h_i=h_i^\ast \land r_i=r_i^\ast]}{N}.
\end{aligned}
\label{eq:attachment-scores}
\end{equation}

For each comparison \(A-B\), we use \(R=10{,}000\) paired sentence-bootstrap samples with seed 42. Each sample draws 2,077 sentences with replacement, uses the same sentence indices for both systems, and recomputes token-weighted LAS using the number of scored tokens in that sample as its denominator. We report pointwise 95\% confidence intervals from the 2.5th and 97.5th percentiles of the sampled differences. We estimate an approximate two-sided percentile-bootstrap p-value as:

\begin{equation}
\begin{aligned}
\widehat{p}_{-} &=
\frac{1+\sum_{b=1}^{R}\mathbf{1}[\Delta_b\leq 0]}{R+1}, \\
\widehat{p}_{+} &=
\frac{1+\sum_{b=1}^{R}\mathbf{1}[\Delta_b\geq 0]}{R+1}, \\
p &= \min\!\left(1,2\min\!\left(\widehat{p}_{-},\widehat{p}_{+}\right)\right).
\end{aligned}
\label{eq:bootstrap-p}
\end{equation}

where \(\Delta_b\) is the LAS difference in bootstrap sample \(b\). The estimates use add-one smoothing and count zero differences in both tails.

\subsection{Pairwise Comparisons}
\label{app:paired-comparisons}

Table~\ref{tab:appendix-comparisons} gives all 15 comparisons per model. Differences are LAS percentage points for the first condition minus the second. ``Holm (15)'' adjusts all 15 tests separately within each model; it differs from the joint four-test correction for the focal comparisons in Section~\ref{sec:metrics}. ``Raw p'' is the approximate bootstrap p-value defined in Appendix~\ref{app:metric-definitions}. Confidence intervals are pointwise, not adjusted for multiple comparisons.

\begin{table*}[t]
\centering
\footnotesize
\setlength{\tabcolsep}{4.5pt}
\textbf{(a) gpt-oss}\\[2pt]
\begin{tabular}{llrrrr}
\toprule
\textbf{Comparison (first $-$ second)} & \textbf{Status} & \textbf{$\Delta$LAS (pp)} & \textbf{95\% CI} & \textbf{Raw \(p\)} & \textbf{Holm \(p\) (15)} \\
\midrule
Zero-Shot $-$ Fixed Few-Shot & Exploratory & $-2.65$ & [$-3.49$, $-1.81$] & .0002 & .0030 \\
Zero-Shot $-$ POS-Signature Few-Shot & Exploratory & $-7.74$ & [$-8.62$, $-6.86$] & .0002 & .0030 \\
Zero-Shot $-$ Semantic Few-Shot & Exploratory & $-5.87$ & [$-6.73$, $-4.99$] & .0002 & .0030 \\
Zero-Shot $-$ Chain-of-Thought & Exploratory & $-5.30$ & [$-6.19$, $-4.41$] & .0002 & .0030 \\
Zero-Shot $-$ Critique-Refine & Exploratory & $-5.74$ & [$-6.57$, $-4.87$] & .0002 & .0030 \\
Fixed Few-Shot $-$ POS-Signature Few-Shot & Focal & $-5.09$ & [$-5.98$, $-4.21$] & .0002 & .0030 \\
Fixed Few-Shot $-$ Semantic Few-Shot & Exploratory & $-3.22$ & [$-4.07$, $-2.37$] & .0002 & .0030 \\
Fixed Few-Shot $-$ Chain-of-Thought & Exploratory & $-2.65$ & [$-3.43$, $-1.83$] & .0002 & .0030 \\
Fixed Few-Shot $-$ Critique-Refine & Exploratory & $-3.09$ & [$-3.84$, $-2.32$] & .0002 & .0030 \\
POS-Signature Few-Shot $-$ Semantic Few-Shot & Focal & $+1.87$ & [$+1.06$, $+2.69$] & .0002 & .0030 \\
POS-Signature Few-Shot $-$ Chain-of-Thought & Exploratory & $+2.44$ & [$+1.54$, $+3.34$] & .0002 & .0030 \\
POS-Signature Few-Shot $-$ Critique-Refine & Exploratory & $+2.00$ & [$+1.12$, $+2.89$] & .0002 & .0030 \\
Semantic Few-Shot $-$ Chain-of-Thought & Exploratory & $+0.57$ & [$-0.33$, $+1.48$] & .2060 & .4120 \\
Semantic Few-Shot $-$ Critique-Refine & Exploratory & $+0.13$ & [$-0.72$, $+0.99$] & .7599 & .7599 \\
Chain-of-Thought $-$ Critique-Refine & Exploratory & $-0.44$ & [$-0.91$, $+0.02$] & .0634 & .1902 \\
\bottomrule
\end{tabular}\\[10pt]
\textbf{(b) Qwen}\\[2pt]
\begin{tabular}{llrrrr}
\toprule
\textbf{Comparison (first $-$ second)} & \textbf{Status} & \textbf{$\Delta$LAS (pp)} & \textbf{95\% CI} & \textbf{Raw \(p\)} & \textbf{Holm \(p\) (15)} \\
\midrule
Zero-Shot $-$ Fixed Few-Shot & Exploratory & $-2.45$ & [$-3.14$, $-1.75$] & .0002 & .0030 \\
Zero-Shot $-$ POS-Signature Few-Shot & Exploratory & $-10.31$ & [$-11.13$, $-9.52$] & .0002 & .0030 \\
Zero-Shot $-$ Semantic Few-Shot & Exploratory & $-7.63$ & [$-8.44$, $-6.85$] & .0002 & .0030 \\
Zero-Shot $-$ Chain-of-Thought & Exploratory & $+1.18$ & [$+0.21$, $+2.22$] & .0158 & .0316 \\
Zero-Shot $-$ Critique-Refine & Exploratory & $-0.72$ & [$-1.64$, $+0.28$] & .1484 & .1484 \\
Fixed Few-Shot $-$ POS-Signature Few-Shot & Focal & $-7.86$ & [$-8.71$, $-7.05$] & .0002 & .0030 \\
Fixed Few-Shot $-$ Semantic Few-Shot & Exploratory & $-5.18$ & [$-6.05$, $-4.35$] & .0002 & .0030 \\
Fixed Few-Shot $-$ Chain-of-Thought & Exploratory & $+3.63$ & [$+2.63$, $+4.69$] & .0002 & .0030 \\
Fixed Few-Shot $-$ Critique-Refine & Exploratory & $+1.73$ & [$+0.77$, $+2.79$] & .0004 & .0030 \\
POS-Signature Few-Shot $-$ Semantic Few-Shot & Focal & $+2.68$ & [$+1.97$, $+3.41$] & .0002 & .0030 \\
POS-Signature Few-Shot $-$ Chain-of-Thought & Exploratory & $+11.49$ & [$+10.45$, $+12.63$] & .0002 & .0030 \\
POS-Signature Few-Shot $-$ Critique-Refine & Exploratory & $+9.59$ & [$+8.56$, $+10.71$] & .0002 & .0030 \\
Semantic Few-Shot $-$ Chain-of-Thought & Exploratory & $+8.81$ & [$+7.78$, $+9.89$] & .0002 & .0030 \\
Semantic Few-Shot $-$ Critique-Refine & Exploratory & $+6.91$ & [$+5.90$, $+7.97$] & .0002 & .0030 \\
Chain-of-Thought $-$ Critique-Refine & Exploratory & $-1.90$ & [$-2.28$, $-1.53$] & .0002 & .0030 \\
\bottomrule
\end{tabular}
\caption{Paired LAS comparisons for both models, using all 2,077 sentences. The two rows marked Focal per model form the joint four-test family; the displayed Holm values instead adjust all 15 comparisons within each model.}
\label{tab:appendix-comparisons}
\end{table*}

The focal rows compare POS-Signature with Fixed or Semantic retrieval. For Qwen, the three absent Critique-Refine outputs remain in the full test-set denominator with zero credit.

\subsection{Results by Dependency Relation}
\label{app:relation-results}

Table~\ref{tab:selected-relations} reports six diagnostic relations. Gold-token counts are shared across conditions; complete results for all 48 scored relation labels are available in \path{results/per_relation.json}. These comparisons are descriptive, without relation-level significance tests.

\begin{table*}[tb]
\centering
\footnotesize
\begin{tabular}{lrrrrr}
\toprule
\textbf{Relation} & \textbf{Gold support} & \multicolumn{2}{c}{\textbf{gpt-oss LAS (\%)}} & \multicolumn{2}{c}{\textbf{Qwen LAS (\%)}} \\
\cmidrule(lr){3-4}\cmidrule(lr){5-6}
 & & \textbf{POS-Signature} & \textbf{Semantic} & \textbf{POS-Signature} & \textbf{Semantic} \\
\midrule
\texttt{nsubj} & 1,950 & 78.4 & 76.4 & 79.8 & 77.4 \\
\texttt{nmod:poss} & 387 & 73.1 & 59.2 & 84.2 & 61.2 \\
\texttt{cc} & 750 & 73.7 & 68.1 & 84.5 & 82.4 \\
\texttt{cop} & 584 & 44.3 & 38.4 & 52.9 & 48.6 \\
\texttt{nmod:unmarked} & 111 & 34.2 & 28.8 & 43.2 & 36.9 \\
\texttt{obl:unmarked} & 113 & 1.8 & 0.0 & 2.7 & 6.2 \\
\bottomrule
\end{tabular}
\caption{LAS (\%) for six diagnostic dependency relations. Gold-token counts are shared across conditions.}
\label{tab:selected-relations}
\end{table*}

POS-Signature improves on Semantic retrieval for several relations, but not uniformly: Qwen performs better with Semantic retrieval on \texttt{obl:unmarked}. The small counts for the two \texttt{unmarked} relations warrant caution.

\subsection{Results by Sentence Length}
\label{app:length-results}

Length is the number of non-punctuation gold tokens. Table~\ref{tab:length-stratified} excludes 31 punctuation-only sentences; failed predictions retain zero credit within each bin. Results are descriptive: no interaction test or simultaneous confidence intervals were specified, and the longest bin contains only 11 sentences.

\begin{table*}[t]
\centering
\footnotesize
\textbf{(a) gpt-oss}\\[2pt]
\begin{tabular}{lrrrrrrrr}
\toprule
\textbf{Bin} & \textbf{Sentences} & \textbf{Tokens} & \textbf{Zero} & \textbf{Fixed} & \textbf{POS-Sig} & \textbf{Semantic} & \textbf{CoT} & \textbf{Critique} \\
\midrule
1--10 & 1,227 & 5,749 & 68.3 & 69.1 & 78.9 & 76.0 & 71.7 & 71.6 \\
11--20 & 533 & 7,821 & 65.1 & 68.2 & 74.2 & 71.1 & 71.4 & 71.3 \\
21--30 & 194 & 4,789 & 64.7 & 66.9 & 69.9 & 68.6 & 70.2 & 70.8 \\
31--50 & 81 & 3,008 & 57.7 & 64.7 & 62.6 & 64.2 & 66.2 & 68.1 \\
51+ & 11 & 631 & 58.3 & 55.5 & 57.1 & 58.2 & 51.3 & 55.9 \\
\bottomrule
\end{tabular}\\[10pt]
\textbf{(b) Qwen}\\[2pt]
\begin{tabular}{lrrrrrrrr}
\toprule
\textbf{Bin} & \textbf{Sentences} & \textbf{Tokens} & \textbf{Zero} & \textbf{Fixed} & \textbf{POS-Sig} & \textbf{Semantic} & \textbf{CoT} & \textbf{Critique} \\
\midrule
1--10 & 1,227 & 5,749 & 61.7 & 64.9 & 78.0 & 74.7 & 62.9 & 63.9 \\
11--20 & 533 & 7,821 & 63.9 & 66.2 & 72.8 & 70.7 & 63.2 & 65.1 \\
21--30 & 194 & 4,789 & 64.1 & 66.8 & 71.5 & 69.2 & 61.9 & 64.6 \\
31--50 & 81 & 3,008 & 61.5 & 62.4 & 69.4 & 66.2 & 59.9 & 62.0 \\
51+ & 11 & 631 & 57.7 & 61.2 & 64.5 & 61.0 & 39.1 & 42.2 \\
\bottomrule
\end{tabular}
\caption{Token-weighted LAS (\%) by non-punctuation sentence length. POS-Sig denotes POS-Signature Few-Shot; Critique denotes Critique-Refine.}
\label{tab:length-stratified}
\end{table*}

For gpt-oss, POS-Signature leads in the two shortest bins, while Critique-Refine is numerically higher in the 21--30 and 31--50 bins. POS-Signature leads all Qwen bins. These patterns do not establish a general advantage for reasoning on longer sentences. Exact counts are preserved in \path{results/length_analysis.json}.

\subsection{Refinement Changes}
\label{app:refinement-edits}

An edit changes a token's predicted head, relation label, or both. A prediction is correct only when both fields match the gold annotation.

\begin{table}[tb]
\centering
\footnotesize
\setlength{\tabcolsep}{3.8pt}
\begin{tabular}{L{3.4cm}rr}
\toprule
\textbf{Critique outcome} & \textbf{gpt-oss} & \textbf{Qwen} \\
\midrule
Analyzed non-punctuation tokens & 21,998 & 21,801 \\
Token prediction changed & 2,260 (10.3\%) & 2,084 (9.6\%) \\
Wrong $\rightarrow$ correct & 703 & 756 \\
Correct $\rightarrow$ wrong & 606 & 338 \\
Wrong $\rightarrow$ different wrong & 951 & 990 \\
\midrule
Net LAS-correct tokens & $+97$ & $+418$ \\
Conditional net change & $+0.44$ pp & $+1.92$ pp \\
Full-corpus LAS change & $+0.44$ pp & $+1.90$ pp \\
\bottomrule
\end{tabular}
\caption{Changes from CoT to Critique-Refine. Qwen transition counts use 21,801 tokens with schema-valid parses at both stages; full-corpus gains use 21,998 tokens.}
\label{tab:refinement-edits}
\end{table}

Qwen's transition counts cover the 2,074 sentences with schema-valid outputs at both stages: 21,801 non-punctuation tokens. Transition percentages and conditional gains use this denominator; full-corpus gains use 21,998 tokens. Both models change similar proportions of predictions, but Qwen introduces fewer new errors. These counts are descriptive; uncertainty for the full-corpus LAS contrast is reported in the main results.

\subsection{Output and Tree Validity}
\label{app:tree-audit}

Valid JSON does not guarantee a valid dependency tree. Table~\ref{tab:output-reliability} distinguishes schema-valid responses, root warnings, and fully tree-conforming outputs. A root warning indicates no token or multiple tokens with \texttt{head=0} in a schema-valid response.

A tree-conforming output has exactly the gold token IDs, legal head IDs, one root, no self-loops, and a path from every token to the root. These checks include punctuation, are separate from UAS/LAS, and do not check projectivity, token-form fidelity, or tag/label inventory membership.

\begin{table*}[t]
\centering
\footnotesize
\begin{tabular}{llrrr}
\toprule
\textbf{Model} & \textbf{Condition} & \textbf{Schema-valid} & \textbf{Root warnings} & \textbf{Tree-conforming} \\
\midrule
gpt-oss & Zero-Shot & 2,077 & 5 & 2,017 \\
gpt-oss & Fixed Few-Shot & 2,077 & 1 & 2,035 \\
gpt-oss & POS-Signature Few-Shot & 2,077 & 1 & 2,005 \\
gpt-oss & Semantic Few-Shot & 2,077 & 3 & 2,009 \\
gpt-oss & Chain-of-Thought & 2,077 & 2 & 2,044 \\
gpt-oss & Critique-Refine & 2,077 & 2 & 2,031 \\
Qwen & Zero-Shot & 2,071 & 12 & 2,017 \\
Qwen & Fixed Few-Shot & 2,076 & 14 & 2,044 \\
Qwen & POS-Signature Few-Shot & 2,076 & 6 & 2,054 \\
Qwen & Semantic Few-Shot & 2,077 & 4 & 2,058 \\
Qwen & Chain-of-Thought & 2,074 & 3 & 2,013 \\
Qwen & Critique-Refine & 2,074 & 5 & 2,028 \\
\bottomrule
\end{tabular}
\caption{Output validity, out of 2,077 sentences per condition. Root warnings count schema-valid responses with no token or multiple tokens having \texttt{head=0}; tree checks impose the additional requirements described in the text.}
\label{tab:output-reliability}
\end{table*}

All gpt-oss responses are schema-valid. For Qwen, failures comprise six malformed Zero-Shot outputs, one malformed Fixed output, one malformed POS-Signature output, and three truncated CoT outputs. The latter prevent three Critique-Refine calls; these absent outputs remain in the denominator. Semantic retrieval has no schema failures. Intermediate token-ID and head-validity counts are retained in the artifact.

\end{document}